\documentclass{article}
\usepackage{graphicx} % Required for inserting images
\usepackage{multirow}
\usepackage{color}
\usepackage{xcolor}
\usepackage{colortbl}
\usepackage{todonotes}
\usepackage{url}
\usepackage{hyperref}
\usepackage{appendix}
\usepackage{authblk}

\definecolor{myred}{rgb}{1,0.78,0.80}
\definecolor{mygreen}{rgb}{0.77,0.93,0.80}
\definecolor{myblue}{rgb}{0.79,0.85,0.97}
\definecolor{myyellow}{rgb}{0.96,0.92,0.69}

\title{Critical-Questions-of-Thought: \\
 Steering LLM reasoning with Argumentative Querying}
\author[1]{Federico Castagna}
\author[2]{Isabel Sassoon}
\author[3]{Simon Parsons}
\affil[1,2]{\small{Brunel University of London, London, UK}}
\affil[3]{\small{University of Lincoln, Lincoln, UK}}
\date{}
\begin{document}

\maketitle
\begin{abstract}
    Studies have underscored how, regardless of the recent breakthrough and swift advances in AI research, even state-of-the-art Large Language models (LLMs) continue to struggle when performing logical and mathematical reasoning. The results seem to suggest that LLMs still work as (highly advanced) data pattern identifiers, scoring poorly when attempting to generalise and solve reasoning problems the models have never previously seen or that are not close to samples presented in their training data. To address this compelling concern, this paper makes use of the notion of \emph{critical questions} from the literature on \emph{argumentation theory}, focusing in particular on Toulmin's model of argumentation. We show that employing these critical questions can improve the reasoning capabilities of LLMs. By probing the rationale behind the models' reasoning process, the LLM can assess whether some logical mistake is occurring and correct it before providing the final reply to the user prompt. The underlying idea is drawn from the gold standard of any valid argumentative procedure: the conclusion is valid if it is entailed by accepted premises. Or, to paraphrase such Aristotelian principle in a real-world approximation, characterised by incomplete information and presumptive logic, the conclusion is valid if not proved otherwise. This approach successfully steers the models' output through a reasoning pipeline, resulting in better performance against the baseline and its Chain-of-Thought (CoT) implementation. To this end, an extensive evaluation of the proposed approach on the MT-Bench Reasoning and Math tasks across a range of LLMs is provided.
\end{abstract}

\section{Introduction}
Despite the remarkable leap forward and recent advancements in AI research, it is a well-known issue that even state-of-the-art Large Language models (LLMs) experience sustained difficulties with logical and mathematical reasoning \cite{srivastava2024functional,mirzadeh2024gsm}. The results seem to suggest that generative AIs still work as (highly advanced) data pattern identifiers (even in text-to-video models \cite{kang2024farvideogenerationworld}), scoring poorly when the input lies in reasoning problems the models have never previously seen or that are not close to samples presented in their training data. Alternatively, it has been postulated that LLMs only memorize a specific set of heuristics without being able to generalize properly when dealing with arithmetic \cite{nikankin2024arithmetic}. This is somehow surprising given the introduction of the test-time compute paradigm, where the allocation of more compute time during the model inference process should ensure better performance \cite{o1,r1,qwq-32b-preview}. This is more important than ever now that we are starting to see some diminishing returns from the previous pre-training scaling law paradigm, i.e., the idea that increasing training data and model parameters will improve overall performance \cite{reuters,information,bloomberg}. To address this compelling concern, we harness and expand the promising test-time compute approach (proved to satisfy a scaling law similar to the previous pre-training hypothesis \cite{chen2024simple}) by employing the notion of critical questions from argumentation theory to improve the reasoning capabilities of LLMs. Harnessing computational argumentation to enhance LLMs' thinking processes is briefly explored in the literature and hints at better performance achievements \cite{castagna2024can,castagna2024computational}. However, unlike previous approaches, what critical questions provide is a mechanism for probing the reasoning process, thus checking the rationale behind the inferences drawn by the models. Given such examination, the LLM is able to assess whether some logical mistake is occurring and can correct it before providing the final reply to the user prompt. The underlying idea is drawn from the gold standard of any valid argumentative procedure: the conclusion is valid if it is entailed by accepted premises \cite{sep-aristotle-logic}. In other words, adapting a form of the Aristotelian principle to a real-world context marked by incomplete information and presumptive reasoning, the conclusion stands as valid unless disproved. In particular, our work sits within the tradition of Toulmin's schema and its defeasible account of argumentative conclusions. 

The main contributions arising from this paper can be summarized as:
\begin{enumerate}
\item The introduction of the novel Critical-Questions-of-Thought (CQoT) approach to support reasoning in LLMs;
\item The provision of an extensive evaluation of the proposed approach on the MT-Bench Reasoning and Math tasks, across a range of LLMs.
\end{enumerate}
Overall, we show that the CQoT technique provides remarkable improvements over the LLMs' baseline performance, thus further corroborating the test-time compute hypothesis: it is possible to 
enhance the capabilities of generative AI models by granting more ``time to think''.   
Note that our work is aligned with the principle of open science, which we advocate for, and hence only makes use of LLMs that are freely available at the time of the research, either as open source or through a limited free plan. Results, evaluation scores and the complete pipeline, rendered as a Python script, can be found at \url{https://github.com/FCast07/CQoT}.

The structure of this paper is as follows. We start by providing an overview of the required theoretical background, focusing on argumentation notions (including schemes and critical questions) and LLMs in Section \ref{sec:background}. We then proceed by specifying the methodology employed and by detailing the CQoT pipeline in Section \ref{sec:methodology}. The evaluation and analysis of the results obtained by testing the CQoT approach are carried out in Section \ref{sec:evaluation}, whereas a further discussion is given in Section \ref{sec:discussion}. Finally, we review the related literature and outline potential future work in Section \ref{sec:future works}, after which we conclude in Section \ref{sec:conclusion}.

\section{Background}
\label{sec:background}
This paper combines several components, each representing an essential building block of the CQoT pipeline introduced herein. In this section, we will unpack the notions and tools underpinning such a pipeline.

\subsection{Argumentation}

The term ``argumentation'', as used in computer science, refers to a wide range of work rooted in philosophy and the study of human reasoning. 
In this tradition, a seminal work is Toulmin's ``The Uses of Argument'' \cite{toulmin:book} which began to formalize the idea that in reasoning it is not just the conclusion that is important, but also the rationale behind the reached conclusion. 
In Toulmin's view, all conclusions are defeasible in the sense that they can be overturned by subsequent inferences.
Conclusions, which Toulmin calls ``claims'', are constructed from \emph{data} about the world, are supported by a
\emph{warrant}, a motive for thinking that the conclusion holds, and this itself is derived from some \emph{backing} (experience or experimental data). 
This whole structure is an \emph{argument}. 
All claims may be subject to a \emph{rebuttal}, and the final truth will only be determined by taking into account all such rebuttals (each of which is itself an argument), examining the warrants, and reaching a verdict on which argument is most plausible. 

\begin{figure}
    \centering
    \includegraphics[width=0.95\linewidth]{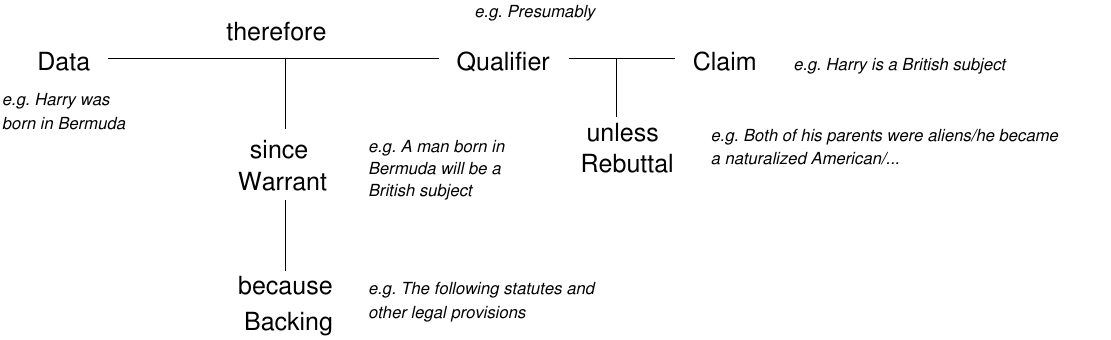}
    \caption{\footnotesize{Toulmin's schema: the case of Harry's nationality.}}
    \label{fig:schema}
\end{figure}

An example of the use of this kind of reasoning, drawn as a formal schema, and taken from \cite{toulmin:book}, is shown in Figure~\ref{fig:schema}. 
Here, we are reasoning about the nationality of an individual ``Harry'', under the rules for British nationality that held at the time that Toulmin was writing.
Here, the data is that Harry was born in Bermuda.
From this it was then possible to make the claim that Harry was British, this conclusion warranted by the fact that, at the time, an individual born in Bermuda was a British citizen.
(This warrant was itself backed by the law of the time concerning British nationality.)
The schema explicitly records that the claim is not certain --- a fact denoted by the attachment of a \emph{qualifier}, such as ``presumably'' --- and the fact that the claim may be subject to a rebuttal, such as the fact that Harry might have given up his British nationality.

We will come back to Toulmin's schema below as a convenient framework to consider how we might bolster the performance of reasoning in LLMs, but before doing that, we should acknowledge other work in what has a significant sub-field within research on artificial intelligence.

One way to consider the kind of reasoning captured by argumentation is as a formal representation that extends classical logic to permit defeasible reasoning.
This is important because it seems that one way of tractably handling the complexity of lots of knowledge about the world is to permit \emph{presumptive} reasoning, that is reasoning that allows for ``jumping to conclusions''. 
(This makes it possible to represent general patterns of reasoning that summarize plenty of specific information.)
However, jumping to conclusions may sometimes lead to incorrect claims, and these then need to be withdrawn, so any systems of presumptive reasoning need to be able to reason defeasibly.

As a result, much of the work on argumentation has been concerned with defeasibility.
The earliest work on argumentation within computer science is due to Loui \cite{loui:ci} and Lin
\cite{lin:shoham:kr89,lin:ci}, the former laying out in broad terms the capabilities of argumentation, while the latter focussed on capturing existing systems of nonmonotonic reasoning (such as default
logic \cite{reiter:ai} and autoepistemic logic \cite{moore:ai}). 
At the same time, Pollock \cite{pollock:cognitive,pollock:ai} was exploring the possibilities of argumentation in great detail, examining the structure of arguments and considering different ways in
which one argument might defeat another.

The seminal work in this area, however, is due to Dung \cite{dung1995acceptability}, who introduced the idea that one could consider the properties of a set of arguments at a purely abstract level. Given a set of arguments
and a relation that identifies pairs of arguments where one attacks the other, it is possible to establish several well-founded principles under which sets of \emph{acceptable} arguments may be determined. For
example, any argument that is not attacked is acceptable, and any argument that is attacked by an acceptable argument is not acceptable (this leads to a simple fixpoint definition of acceptability). This basic idea has been widely adapted and extended, see for example \cite{baroni2018handbook}.

\subsection{Argument Structure \& Critical Questions}

Toulmin's schema provides us with part of what we need to augment LLM reasoning, and we combine that idea with the more recent notions of argument schemes and critical questions \cite{walton1996argumentation,walton:reed:macagno:book}.
Broadly speaking, the idea behind \emph{argument schemes} is that presumptive reasoning operates through the application of a number of schemes that capture common patterns of reasoning. 
Since they support presumptive reasoning, they are not expected to be deductive in the truth-preserving sense that classical logic is.
Rather, they capture rules for drawing inferences that will often provide useful conclusions but will sometimes lead to invalid claims.
In that sense, such schemes are more like the default rules of default logic \cite{reiter:ai}.

One of the key features of argumentation schemes is the list of associated \emph{critical questions} (CQs). 
The role of these questions is to highlight when the scheme may or may not lead to reasonable conclusions.
The premises, claim (or conclusion) supported by the argumentation scheme are presumptive, and a premise or claim is withdrawn unless the CQs posed have been answered successfully. 
In general, CQs can also be used as pointers to counter-arguments that themselves instantiate argument schemes. 
CQs can then be used in two different ways: (i) posing CQs to construct counter-arguments that can themselves instantiate schemes with their own CQs; (ii) posing CQs to challenge an existing argument (e.g. questioning a premise) and which can be responded by constructing a supporting argument.

What we do in this paper is think of the LLM as providing reasoning that can be structured as per a Toulmin schema, and then use pertinent critical questions to check the validity of the presumptive conclusions that the LLM has reached. As described above, Toulmin's model of argument includes the following six components:
\begin{itemize}
\item Claim - The conclusion or main argument being made;
\item Qualifier - Limits or conditions on the claim;
\item Data - The facts, evidence, or premises that support the claim;
\item Warrant - The logical connection or justification that bridges the data and claim;
\item Rebuttal - Acknowledgment of counterarguments or exceptions to the claim;
\item Backing - Additional support or justification for the warrant.

\end{itemize}
We thus introduce critical questions that probe whether these elements hold.
(As we shall see in Section~\ref{sec:methodology}, we introduce eight critical questions, particularly targeting the Data and the Warrant components).

\subsection{LLMs}
Large Language Models (LLMs)\footnote{One could argue that the present state-of-the-art consists of Multimodal Models (MMs), i.e., AI models that accept as input (and provide as output) other media besides language. Given the nature of the task we want to achieve, in the present work, we resort to experimenting with language and textual input/output only. We will thus use the term ``LLMs'' even in those cases where referring to them as MMs might be more accurate.} represent one of the current pinnacles of Deep Learning technology stemming from the transformer architecture \cite{vaswani2017attention}. Since the release of ChatGPT in November 2022 \cite{introducinggpt}, the production and deployment of generative AI systems have soared, resulting in new model announcements almost every week. The competition among proprietary companies, the contribution of the open source community and the huge investments made in the sector over the past two years have certainly yielded impressive advancements in the AI landscape. The main actors of this progress include tech firms such as Anthropic, OpenAI, Google, Meta, NVIDIA and many others. In particular, Anthropic's \emph{Claude 3.5 Sonnet} \cite{claude3_5}, OpenAI's \emph{GPT-4o} \cite{openai2024gpt4ocard} and Google's \emph{Gemini 1.5 pro} \cite{geminiteam2024} have alternated as the top three models (according to a plethora of different evaluation benchmarks) during 2024. However, despite their outstanding performances, these models are still prone to errors, especially when handling tasks that require complex reasoning capabilities.  Some studies claim that LLMs are just proficient at memorizing training data, but still struggle to generalize and undertake tasks which are unknown or unlike anything they have previously seen \cite{srivastava2024functional,mirzadeh2024gsm}.

\subsubsection{Chain-of-Thought}
To improve LLM capabilities and curtail their shortcomings, several different techniques have been developed. Some of such approaches involve training, fine-tuning or other resource-intensive post-processing stages. Researchers also engaged with lightweight methods which, in some cases, proved even more successful than the aforementioned ones. Among these examples, prompting engineering techniques, i.e., practices to effectively craft prompts that optimize models' output, have been largely employed to enhance LLM logical thinking \cite{sahoo2024systematic}. Chain of Thought (CoT), probably the most influential of such techniques, consists of a prompting strategy that details a series of intermediate reasoning steps to achieve better performance in arithmetic, symbolic and commonsense inferences \cite{wei2022chain}. Zero-shot-CoT is a streamlined version of CoT that is task-agnostic and does not require few-shot examples \cite{kojima2022large}. In the following, we will leverage the first prompt of this (originally twofold) approach, rendered by the well-known phrasing \emph{``Let's think step by step''}. For simplicity, we will refer to it just as CoT.

\subsection{Benchmarks}
In order to assess the overall capabilities of an LLM, thus ranking the best-performing ones, AI researchers needed to design solutions that were more practical than relying on expensive and time-consuming human testing. With this in mind, multiple benchmarks were created, each focusing on evaluating specific models' skill sets \cite{chang2023survey}. For example, MT-Bench is a comprehensive multi-turn benchmark that presents 80 challenging queries, divided into two sub-questions each, covering 8 different domains (writing, roleplay, reasoning, math, coding, extraction, stem, humanities) and LLMs' skills (e.g., causal relations detection, thinking process, creativity, factual knowledge, instructions compliance) \cite{zheng2023judging}. Unlike other popular benchmarks such as Big-Bench Hard \cite{suzgun2022challenging}, MMLU \cite{hendrycks2020measuring}, or GPQA \cite{rein2023gpqa}, it does not present multiple options to choose from nor it (always) includes a fully-fledged reference answer. This allows for a more thorough test of the models' capabilities in an open-ended setting. Given the specific gap we would like to address, we will focus only on the reasoning and math questions included in the MT-Bench.

\section{Methodology}
\label{sec:methodology}

\begin{figure}[t]
    \centering
    \includegraphics[width=1.05\linewidth]{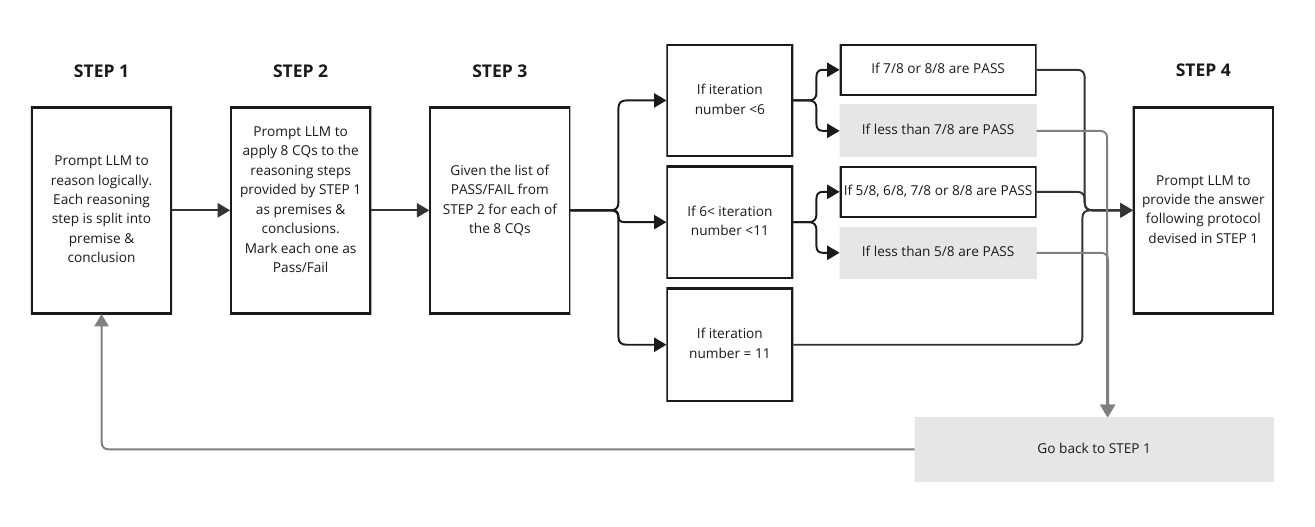}
    \caption{\footnotesize{The four-step process of the CQoT pipeline.}}
    \label{fig:pipeline}
\end{figure}

Our Critical-Questions-of-Thought (CQoT) approach can be summarised by the four steps illustrated in Figure~\ref{fig:pipeline}. The chosen name already hints at the main component of this technique, i.e., the critical questions, whose contribution occur  
in Step 2 of the pipeline. For each CQ we identify which elements of Toulmin's model the question targets.
Note that we are not trying to claim that an LLM follows the Toulmin model of argumentation, but rather that LLMs engage in presumptive reasoning, and so a set of CQs that cover all the elements of Toulmin's general schema of presumptive reasoning provide a suitable battery of queries with which to test any conclusions that an LLM comes up with.
The leveraged critical questions are: 
\begin{enumerate}
\item Does the reasoning process start with clearly defined premises? (Data)

\item Are the premises supported by evidence or accepted facts? (Data)

\item Does the reasoning process use logical connections between premises and conclusions? (Warrant)

\item Are the logical connections used in the reasoning process valid? (Warrant)

\item Does the reasoning process avoid fallacies or logical errors? (Warrant, Backing)

\item Is the conclusion logically derived from the premises? (Claim)

\item Is the reasoning process consistent with established knowledge or principles? (Backing)

\item Does the reasoning process lead to a conclusion that is plausible and reasonable? (Claim, Qualifier, Rebuttal)
\end{enumerate}
The reasoning is critically evaluated against these questions, with a conclusion being ruled out unless a majority of the CQs are answered positively.

\noindent Aligned with the open science principle, we tested our approach by resorting only to freely available LLMs, either open source (e.g., hosted on \href{https://huggingface.co}{HuggingFace}) or through a free plan. In particular, we chose \emph{Gemini 1.5-pro-001} due to its being among the top-performing ones at the time of the experiments, and we had access to it thanks to the free credits provided by Google Cloud. Similarly, given their high performances and the availability of a free plan, we also picked \emph{GPT-4o} and \emph{Claude Sonnet 3.5} (both October 2024 versions). Finally, we opted for \emph{Llama 3.1-70b-Instruct} and \emph{Nemotron-51b-Instruct} because, once again, they were among the top-tier open-sourced LLMs for their size category at the time of the experiment. We also tried to diversify the models as much as possible by selecting the products of different companies to test our approach on a broader spectrum of samples.

\subsection{Critical-Questions-of-Thought pipeline}
\begin{figure}[h!]
    \centering
    %\hspace{-2cm}
\includegraphics[width=1\linewidth]{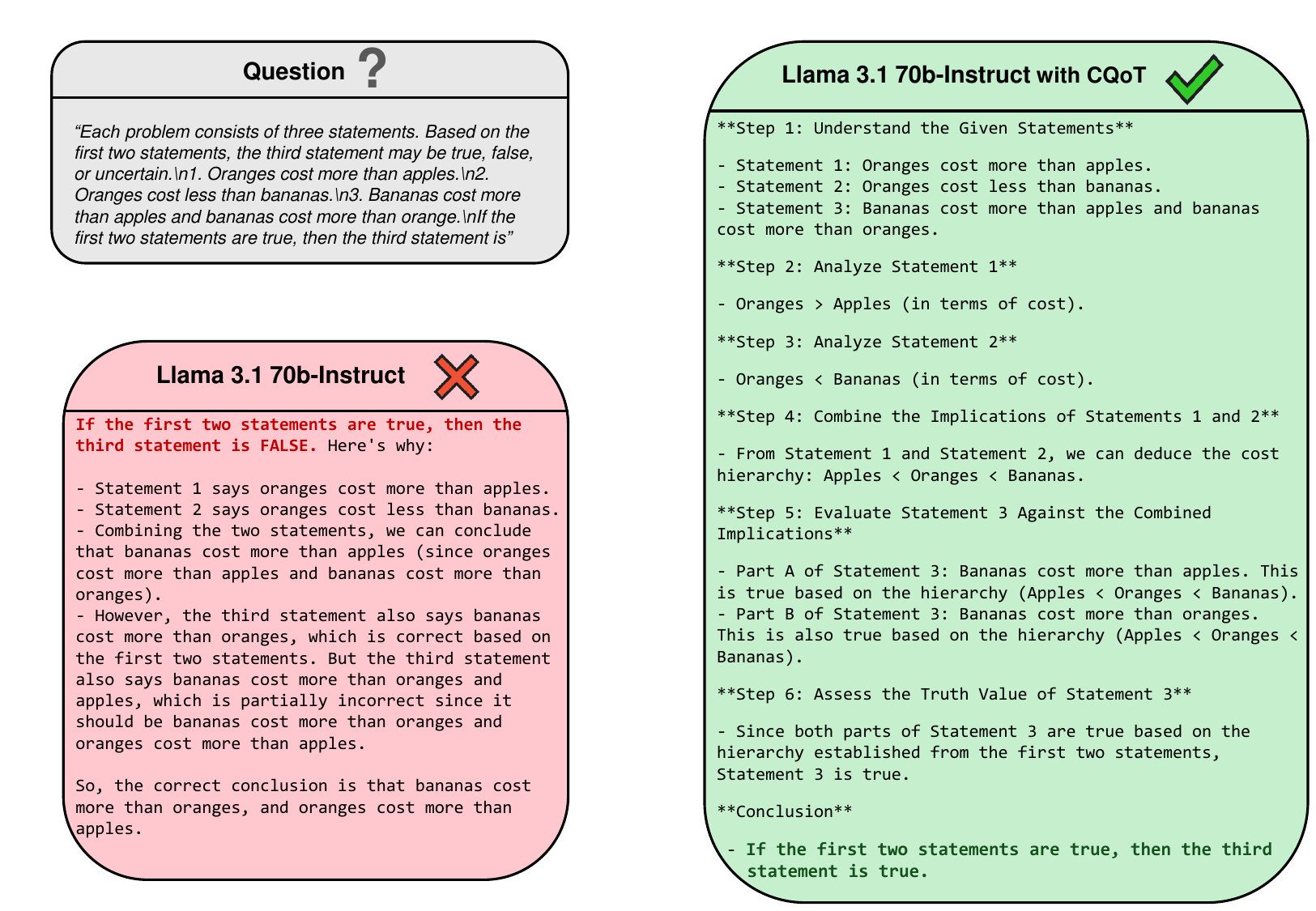}
%\vspace{1cm}
    \caption{\footnotesize{Comparison between the responses given by the baseline Llama 3.1 70b-Instruct (wrong, red coloured) and its CQoT counterpart (correct, green coloured). Notice that the multiple `Step' mentioned in the latter reply do not refer to the CQoT pipeline: it is just how the output has been phrased by the model.}}
    \label{fig:CQoT_examples}
    \end{figure}

Critical-Questions-of-Thought is a simple approach that can be implemented on top of any LLMs. In this section, we introduce and break down each of the four stages that compose the CQoT pipeline, as depicted in Figure \ref{fig:pipeline}. In a nutshell, Steps 1, 2 and 4 can be interpreted as different prompting strategies leveraging the outcome of the previous steps. Step 3 operates as a checkpoint. It verifies the model's responses to the CQs and decides whether the pipeline should continue to the final phase or iterate over, effectively starting again from the initial stage. More precisely:
\vspace{0.2cm}
\begin{itemize}
    \item \textbf{\fbox{Step 1.}} Prompt the LLM to reason logically about the input query and sketch a plan to follow. The model must divide each reasoning process into premises and conclusions so that each conclusion derives from the respective premises. No final response should be provided.
    \item \textbf{\fbox{Step 2.}} The model will now have to assess the validity of the previously envisaged reasoning steps by checking whether they address the critical questions devised in Section \ref{sec:methodology}. All such CQs are related to the validity of logical reasoning and refer to specific elements of the Toulmin representation of arguments.
    \item \textbf{\fbox{Step 3.}} The model has to evaluate the replies to the critical questions. If there are at least 7/8 positive answers, then the pipeline will move to \textbf{Step 4}. Otherwise, the pipeline will start again from \textbf{Step 1}. This will occur for a maximum of five iterations, after which the requirement will drop to a minimum of 5/8 positive answers. If this score has not been reached after a total of ten iterations, the pipeline will, regardless, move to \textbf{Step 4} with the latest generated reasoning plan. 
    \item \textbf{\fbox{Step 4.}} Having determined the correctness of the model reasoning process, the LLM is now required to strictly follow it in order to provide the final reply to the original user prompt.
\end{itemize}
\vspace{0.2cm}
CQoT proceeding has been automated as a Python class, and we publicly share the code on our \href{https://github.com/FCast07/CQoT}{GitHub repo}. Notice that, in order to use the script, we need to have access to the underlying LLM weights. Alternatively, our approach can also be leveraged by proprietary models by manually following the four-step pipeline and their specific prompts (see Figure \ref{fig:CQoT_prompts} in the Appendix for in-depth instructions), where the user will act as the `verifier' (Step 3).
As a comparative example of the results achieved by the CQoT approach against the baseline model, we can examine Figure \ref{fig:CQoT_examples}. Here, we can see how Llama 3.1 70b-Instruct, when challenged by a logical puzzle, outputs the wrong response. However, when leveraging the CQoT pipeline (hidden for simplicity in this instance), the LLM answers correctly. More precisely, given the riddle: 
\begin{quote}
   \emph{``Each problem consists of three statements. Based on the first two statements, the third statement may be true, false, or uncertain. n1. Oranges cost more than apples. n2. Oranges cost less than bananas. n3. Bananas cost more than apples and bananas cost more than orange. If the first two statements are true, then the third statement is"} 
\end{quote}
the baseline model replied \emph{``If the first two statements are true, then the
third statement is FALSE."}, whereas the CQoT augmented counterpart rightly responded \emph{``If the first two statements are true, then the third statement is true."}.

\section{Evaluation}
\label{sec:evaluation}
\subsection{Experimental setup}
In this section, we lay out the specifics of how we appraised the CQoT pipeline (Figure \ref{fig:pipeline}) against the chosen LLMs in their standard and CoT-augmented versions. We also tested the same baseline models by means of an ablation study. Through a slightly modified version of the previously outlined CQoT pipeline obtained by skipping the second and third stages (Figure \ref{fig:ablation pipeline}), we checked whether the addition provided by the CQs is what increases the LLMs' reasoning performances. Our findings showed that the presence of CQs indeed enhances the models' ability to reply logically. Once again, we share all the evaluation results, including LLMs' responses, judge scores and their mean, on our \href{https://github.com/FCast07/CQoT}{GitHub repo}.

\subsubsection{Parameters}
We previously reported that the experiment has been conducted with both open source and (free plan version of) proprietary LLMs. Where we had access to the full models' inference parameters, i.e., with Llama 3.1-70b-Instruct and Nemotron-51b-Instruct, we tuned them as follows. 
\texttt{return\_full\_text:}False; \texttt{temperature:$0.8$} (\texttt{temperature:$0.2$} for Step 2 and Step 4, which need to be as objective as possible); \texttt{top\_p:$0.95$}; \texttt{do\_sample:}True; \texttt{max\_new\_tokens:$2000$}. The underlying idea of this setting was to control the creativity of the LLM while saving the available compute. When leveraging Gemini 1.5-pro-001 on the Google Vertex AI platform, we were able to adjust fewer parameters: \texttt{temperature:$1$} (\texttt{temperature:$0$} for Step 2 and Step 4, which need to be as objective as possible); 
 \texttt{top\_p:$0.95$}; \texttt{max\_new\_tokens:$8192$}. The rationale behind the latter value mostly stems from the presence of fast computing resources on the platform. 

\subsubsection{MT-Bench} 
As mentioned earlier, we opted for MT-Bench math and reasoning questions\footnote{We closely followed the MT-Bench prompts as stored on  \href{https://huggingface.co/datasets/HuggingFaceH4/mt_bench_prompts}{HuggingFace}.} to assess the performances of our pipeline. Unfortunately, it is a well-known issue that popular benchmark data may contaminate the models' training data, thus affecting the validity of the LLMs' output. This problem has led the AI community to devise alternative evaluation methods such as LiveBench, which contains frequently updated questions from recent information sources \cite{white2024livebench}, and Chatbot Arena, an open platform to appraise LLMs via human preferences \cite{chiang2024chatbot}. Although MT-Bench has been available for about a year at the time of the writing \cite{zheng2023judging} and the risk that the models' training data has been contaminated with MT-Bench prompt exists, we tested the same LLMs with and without CoT and CQoT, thus ensuring a fair comparison of the performances (i.e., if the baseline has been contaminated, so is the CoT and the CQoT version, therefore effectively putting them on equal ground for evaluation)\footnote{Indeed, using MT-Bench, in this case, provides a stiffer test for the CQoT approach than using a benchmark that we know has not been contaminated. If there is contamination, it will tend to raise the performance of the LLMs on MT-Bench, reducing the headroom for CQoT to improve their performance.}. For the CoT-augmented version of the underlying model, we employed a specific prompt, as detailed in Figure \ref{fig:CoT_prompt}. Overall, each harnessed model has been input with a total of 40 questions, evenly divided between reasoning and math topics. We argue that the collected answers suffice to render a meaningful sample of the differences between the standard LLM, its CoT implementation, and the CQoT pipeline. %To enrich the comparison, we also included the CoT augmented version of the underlying model baseline in the test against the MT-Bench selected tasks.   

\subsubsection{Ablation study}
\begin{figure}[t]
    \centering
    \includegraphics[width=0.8\linewidth]{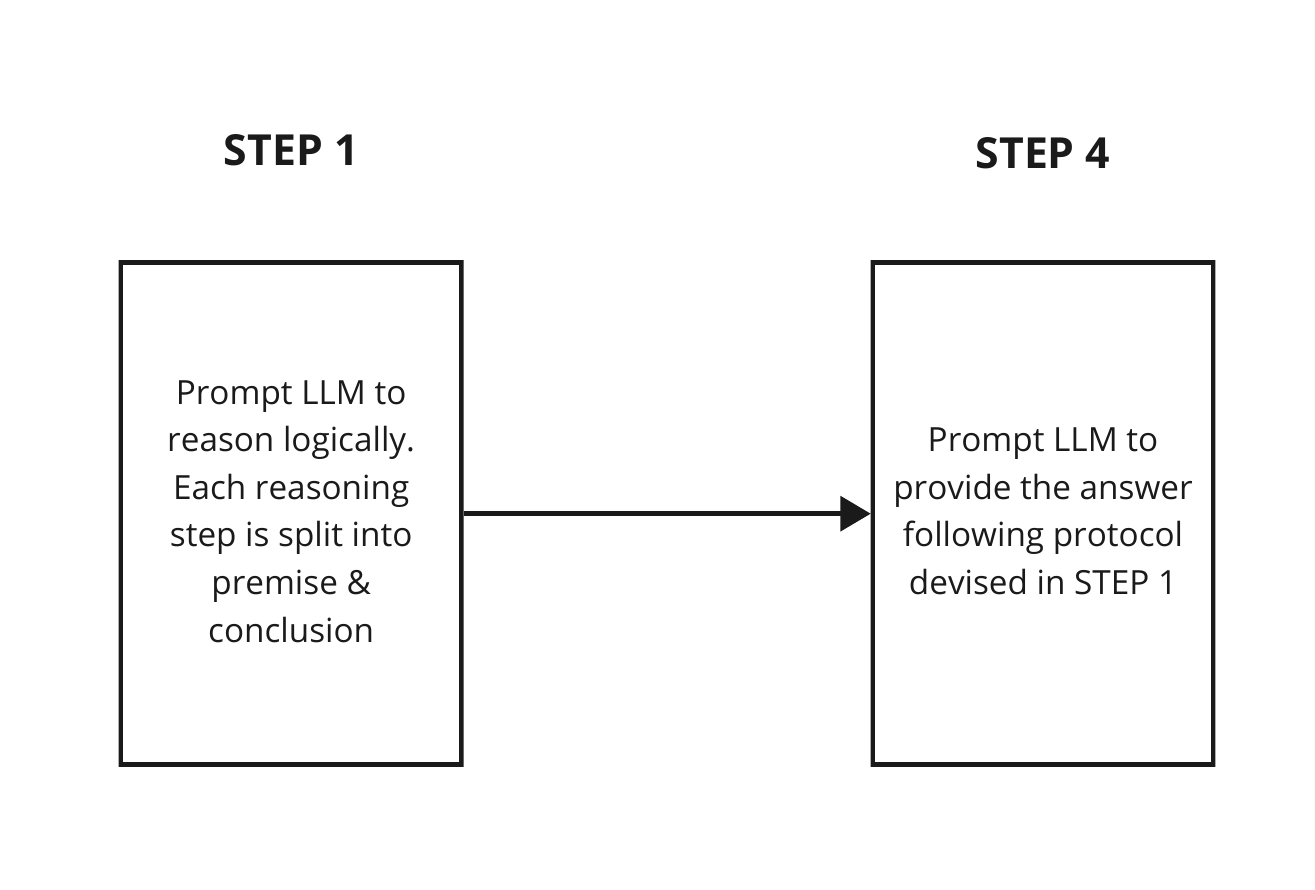}
    \caption{\footnotesize{Two-step pipeline for the ablation study.}}
    \label{fig:ablation pipeline}
\end{figure}

To effectively verify that the CQoT results were indeed dependent on the probing impact of the critical questions, we devised an additional test. We evaluated both Llama 3.1 70b-Instruct and Nemotron-51b-Instruct on a slightly different pipeline composed only of the first and fourth steps of the CQoT method. We opted for these two models out of the five we previously examined due to their being open-source and thus being easier to use for reproducing our findings. The idea was to determine whether the strengths of the CQoT approach lay solely in the envisaged reasoning plan or whether the presence of CQs was granting further improvements. The full pipeline leveraged for this ablation study is described in Figure \ref{fig:ablation pipeline}, whereas the respective prompts are presented in Figure \ref{fig:CQoT_onlyreasoning_prompts} in the Appendix.

\subsubsection{LLM as a Judge}
Drawing from the research of Zheng et al. \cite{zheng2023judging}, where it has been established how state-of-the-art LLMs can achieve an agreement rate on par with human experts when assessing model responses, we followed a similar evaluation strategy. When choosing the judge, we considered both GPT-4o and Claude 3.5 Sonnet as promising options. The former was better at appraising math questions, and the latter was stronger at assessing reasoning queries \cite{tan2024judgebench}. Given that, at the time of the experiment, the free tier version of GPT-4o (October 2024 version) was more generous than Claude in granting a larger number of requests per day, we selected the first as the judging model.
The prompts provided to evaluate the LLMs' response quality are also taken from the study conducted by Zheng et al. \cite{zheng2023judging}, although we slightly reshaped such prompts to better reflect the task of handling only reasoning and math-related queries (see Figure \ref{fig:Judge_prompts} in the Appendix). Such prompts instruct the judging model to assess responses using a 1-10 scoring system, where higher grades mean better performances. To avoid fluctuations in the judge rates and restrict the potential randomness of the outcome, we manually prompted GPT-4o multiple times for each answer to be evaluated until the same score occurred three times. Such a score was then identified as being representative of the response of the appraised model. We argue that this method ensures a more reliable grading of the evaluated answers compared to single scoring. Additionally, there were instances where we further prompted the judge with follow-up inputs. In particular, we employed a specific instruction (see Figure \ref{fig:Judge_followup_prompt} in the Appendix) that proved useful in balancing the assessment whenever the score penalized the lack of conciseness (a criterion, we believe, is not fundamental when testing reasoning capabilities). We also repeated the reference answer when the judge wasn't properly acknowledging it (although this was an event that seldom occurred).  

\subsection{Results}

\begin{figure}[h!]
    %\centering
    \hspace{-2cm}
\includegraphics[width=1.3\linewidth]{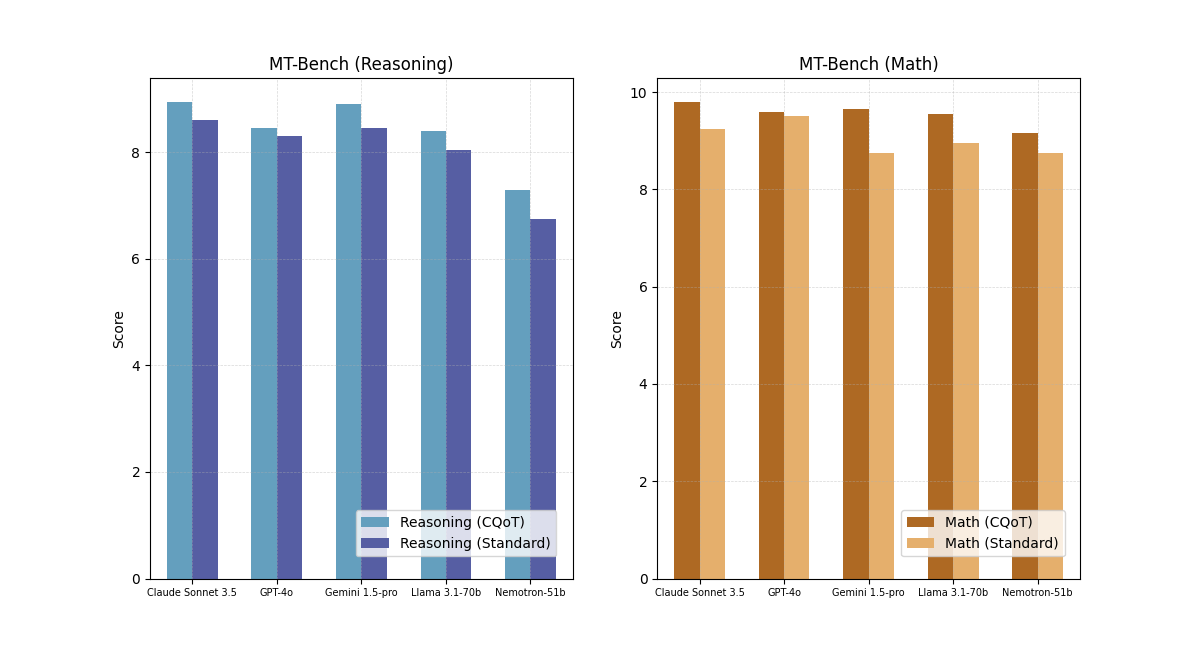}
\vspace{-1cm}
    \caption{\footnotesize{Comparison between performance achieved by the baseline model with (CQoT) and without (Standard) the Critical-Questions-of-Thought approach.}}
    \label{fig:barplot}
    \end{figure}

\begin{figure}[h!]
    \centering
    %\hspace{-1.5cm}
\includegraphics[width=1\linewidth]{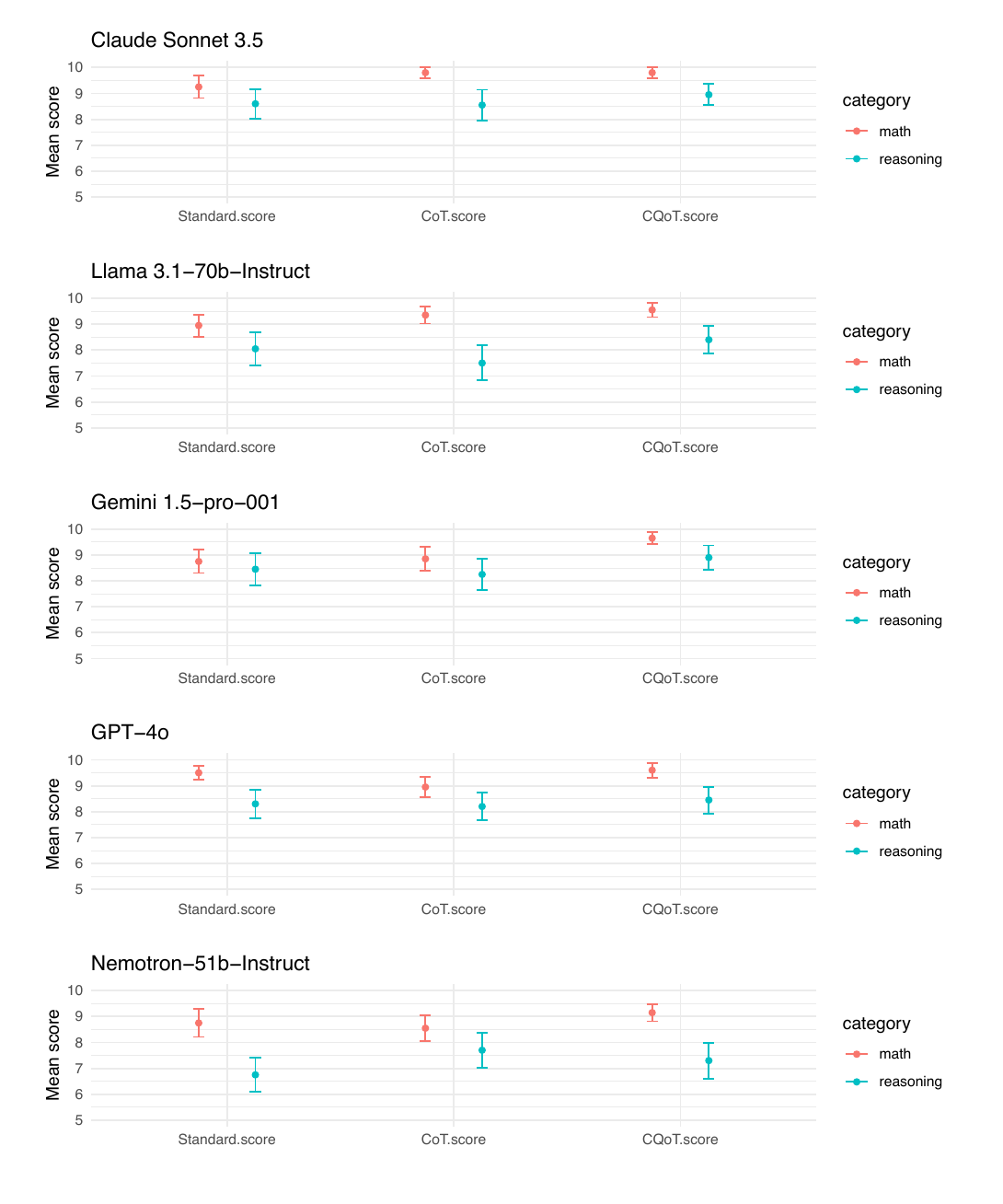}
\vspace{-1cm}
    \caption{\footnotesize{Comparison between performance achieved by the baseline model (Standard), its CoT augmented version (CoT) and the Critical-Questions-of-Thought approach (CQoT). The point represents the mean score, and the bars are the standard error.}}
    \label{fig:boxplot}
    \end{figure}

Results of the laid-out evaluation, obtained by testing the whole pipeline, can be seen in Tables~\ref{tab:eval_scores}, Figure~\ref{fig:barplot} and Figure~\ref{fig:boxplot}, whereas the outcome of the follow-up ablation experiment, comparing a pipeline composed only by Step 1 and 4 against the baseline and its CQoT version, can be seen in Table~\ref{tab:eval_step1_only}. Both tables are arranged in a similar way, emphasizing in bold the LLMs' highest performance according to the task (math or logical reasoning) and the technique harnessed (standard baseline, Step1\_4, CoT or CQoT). The numbers displayed are the mean of the values scored (i.e., assigned by the LLM judge) by a model under a task with a specific technique. Figure~\ref{fig:barplot} graphically showcases the same average numbers by means of a barplot, focusing on the baseline and CQoT, separating the two tasks of comparison. Figure~\ref{fig:boxplot} provides a more detailed visualization of the mean scores (points) per model per task and the respective standard errors (bars). The collected data is examined in the next section.

\begin{table}[h!]
    \centering
    \begin{tabular}{|c|c|c|c|c|c|c|}
         \hline
          \multirow{2}{*}{\textbf{\large Models}}& \multicolumn{3}{c|}{\textbf{\small{MT-Bench (Reasoning)}}}  & \multicolumn{3}{c|}{\textbf{\small{MT-Bench (Math)}}}\\
          \cline{2-7}
           & \textbf{\textit{\footnotesize{Standard}}} & \textbf{\textit{\footnotesize{CoT}}} & \textbf{\textit{\footnotesize{CQoT}}} & \textbf{\textit{\footnotesize{Standard}}} & \textbf{\textit{\footnotesize{CoT}}} & \textbf{\textit{\footnotesize{CQoT}}}\\
         \hline
         \textit{\footnotesize{Claude Sonnet 3.5}} & 8.6 &8.55 &\textbf{8.95} &9.25 & \textbf{9.8} & \textbf{9.8}\\
         \hline
         \textit{\footnotesize{GPT-4o}} & 8.3  &8.2 & \textbf{8.45}&9.5 &8.95 &\textbf{9.6}\\
         \hline
         \textit{\footnotesize{Gemini 1.5-pro-001}}& 8.45 & 8.25 & \textbf{8.9} &8.75 &8.85 & \textbf{9.65}\\
         \hline
         \textit{\footnotesize{Llama 3.1-70b-Instruct}}&8.05 &7.5 &\textbf{8.4} &8.95 &9.35 &\textbf{9.55}\\
         \hline
         \textit{\footnotesize{Nemotron-51b-Instruct}}& 6.75 & \textbf{7.7}& 7.3 &8.75 &8.55 &\textbf{9.15}\\
         \hline
    \end{tabular}
    \caption{\footnotesize{Evaluation scores (mean) in a range of 1-10, with the highest results per row highlighted in bold.}}
    \label{tab:eval_scores}
\end{table}

\begin{table}[h!]
    \centering
    \begin{tabular}{|c|c|c|c|c|c|c|}
         \hline
          \multirow{2}{*}{\textbf{\large Models}}& \multicolumn{3}{c|}{\textbf{\small{MT-Bench (Reasoning)}}}  & \multicolumn{3}{c|}{\textbf{\small{MT-Bench (Math)}}}\\
          \cline{2-7}
           & \textbf{\textit{\footnotesize{Standard}}} & \textbf{\textit{\footnotesize{Step1\_4}}} &\textbf{\textit{\footnotesize{CQoT}}} &
           \textbf{\textit{\footnotesize{Standard}}} & \textbf{\textit{\footnotesize{Step1\_4}}} &  \textbf{\textit{\footnotesize{CQoT}}}\\
         \hline
         \textit{\footnotesize{Llama 3.1-70b-Instruct}}& 8.05& 8.3 &\textbf{8.4} & 8.95 & 9.25 &\textbf{9.55} \\
         \hline
         \textit{\footnotesize{Nemotron-51b-Instruct}}&6.75 &6.85 &\textbf{7.3} &8.75 &9 &\textbf{9.15} \\
         \hline
    \end{tabular}
    \caption{\footnotesize{Evaluation scores (mean) in a range of 1-10, with the highest results per row highlighted in bold. The table compares the outcomes ensuing from the baseline and the full CQoT pipeline with the results achieved by following Step 1 and Step 4 only.}}
    \label{tab:eval_step1_only}
\end{table}

\subsection{Analysis}

Employing the CQoT pipeline creates a significant increase in the performance of models on the MT-Bench Reasoning and Math tasks when compared with both the baseline performance (the raw LLM output) and the performance of the LLMs augmented with CoT. These improvements are particularly marked for GPT-4o, Gemini 1.5-pro-001 and Llama 3.1-70b-Instruct. They are somewhat less impressive for Claude Sonnet 3.5, where CoT and CQoT give the same performance on the Math benchmark, and for Nemotron-51b-Instruct, where CQoT performs worse than CoT on the Reasoning benchmark. However, considering these as 20 separate tests (CQoT against the baseline and against CoT on two domains and across five models), CQoT has the undoubted best performance on 18 of these separate tests and has the joint best performance on another (Table \ref{tab:eval_scores}). The magnitude of the overall improvement can be appreciated in Figure \ref{fig:boxplot}, where CQoT exhibits a lower standard error for the majority of the samples considered, thus showing more stability than the standard and CoT approach. Furthermore, we can quantify the enhancement achieved by CQoT by expressing its results as a percentage, as shown in Table \ref{tab:eval_percentages}. The four columns represent the difference between the mean scores obtained by CQoT with the baseline and the CoT mean scores, respectively. The average of such values is $\mathbf{+4.61\%}$ for reasoning questions and $\mathbf{+5.4\%}$ for math questions, meaning that our technique ensures an overarching improvement of approximately $\mathbf{5\%}$ against the other contending approaches. Finally, notice how, thanks to CQoT, the open source Llama 3.1-70b-Instruct outperforms the baseline responses of the proprietary GPT-4o (also rumoured to be a much larger model) on the MT-Bench Reasoning tasks whilst exceeding all the proprietary LLMs on the MT-Bench Math queries. Similarly, Nemotron-51b-Instruct, which is an even smaller model, 
exceptionally manages to score higher than Gemini 1.5-pro-001 on mathematical questions. As a last remark, we observed how the role of critical questions in probing the reasoning plan sketched by the underlying model is essential in steering the final output towards better (more correct) responses. Indeed, results from the ablation study (Table \ref{tab:eval_step1_only}) show how CQs describe the rationale behind the performance enhancement. Nonetheless, it is interesting to see that a pipeline composed only of Steps 1 and 4 would still score higher than the standard baseline, emphasising how each stage of the CQoT approach contributes to the LLM's improved output. 
\begin{table}[h!]
    \centering
    \begin{tabular}{|c|c|c|c|c|}
         \hline
          \multirow{2}{*}{\textbf{\large Models} \textbf{+ CQoT}}& \multicolumn{2}{c|}{\textbf{\small{MT-Bench (Reasoning)}}}  & \multicolumn{2}{c|}{\textbf{\small{MT-Bench (Math)}}}\\
          \cline{2-5}
           & \textbf{\textit{\footnotesize{Standard}}} & \textbf{\textit{\footnotesize{CoT}}} & \textbf{\textit{\footnotesize{Standard}}} & \textbf{\textit{\footnotesize{CoT}}}\\
         \hline
         \textit{\footnotesize{Claude Sonnet 3.5}} &\cellcolor{mygreen}+4.06\% &\cellcolor{mygreen}+4.68\% &\cellcolor{mygreen}+5.95\% & \cellcolor{myyellow}+0\% \\
         \hline
         \textit{\footnotesize{GPT-4o}} &\cellcolor{mygreen}+1.81\%  &\cellcolor{mygreen}+3.04\% &\cellcolor{mygreen}+1.05\% &\cellcolor{mygreen}+7.26\%\\
         \hline
         \textit{\footnotesize{Gemini 1.5-pro-001}}&\cellcolor{mygreen}+5.33\% &\cellcolor{mygreen}+7.88\% &\cellcolor{mygreen}+10.29\% &\cellcolor{mygreen}+9.04\%\\
         \hline
         \textit{\footnotesize{Llama 3.1-70b-Instruct}}&\cellcolor{mygreen}+4.35\% &\cellcolor{mygreen}+12\% &\cellcolor{mygreen}+6.70\% &\cellcolor{mygreen}+2.14\%\\
         \hline
         \textit{\footnotesize{Nemotron-51b-Instruct}}&\cellcolor{mygreen}+8.15\% &\cellcolor{myred}-5.19\% &\cellcolor{mygreen}+4.57\% &\cellcolor{mygreen}+7.02\% \\
         \hline \cellcolor{myblue}\textit{\small{\textbf{Average}}} &\cellcolor{myblue} \textbf{+4.74\%}&\cellcolor{myblue}\textbf{+4.48\%} &\cellcolor{myblue}\textbf{+5.71\%} &\cellcolor{myblue}\textbf{+5.09\%}\\
         \hline
    \end{tabular}
    \caption{\footnotesize{Percentage of CQoT performances listed in Table \ref{tab:eval_scores}. The columns `Standard' display the score obtained by confronting CQoT with the baseline model results. `CoT' columns showcase the score obtained by comparing CQoT with the baseline model augmented with CoT.}}
    \label{tab:eval_percentages}
\end{table}

\section{Discussion}
\label{sec:discussion}
As humans take more time to elaborate and solve hard problems \cite{kahneman2011thinking}, it is plausible to assume that, by giving more `time to think' to LLMs during inference, the models' capability to correctly address challenging reasoning issues should likewise increase. This is exactly what the test-time compute paradigm has confirmed, especially through LLMs such as OpenAI's o1 \cite{o1}, DeepSeek's R1-Lite-Preview \cite{r1} and Qwen's QwQ \cite{qwq-32b-preview}. However, another main feature of human thinking is its argumentative nature and how it is fundamental to formulate arguments that challenge (or support) specific information to consolidate our knowledge. Indeed, Mercier and Sperber claim that reasoning evolved as a byproduct of the generation and evaluation of arguments \cite{mercier2011humans}. This notion underpins the purpose of the critical questions as means to probe the reliability of the information embedded in the inquired argument. The CQoT approach combines together CQs, inspired by Toulmin's model, and the test-time compute paradigm into a pipeline that allows LLMs to enhance their logical and mathematical skills.  

We noticed that the driving engine of the CQoT pipeline is the reasoning protocol generated in Step 1, which steers the underlying model's thinking process. The LLM's reply will be reliant on such a protocol, thus possibly leading to an erroneous response if the provided reasoning plan presents any mistakes. The employment of CQs (implemented in Step 2 and verified in Step 3) will mostly filter out each of such wrongly designed protocols, but some may still be able to elude the test, especially for smaller and less performing models\footnote{Notice that we will not be able to address this issue by modifying Step 3 and requesting, when the replies to the CQs are not all positive, to output the reasoning protocol that occurred the most during the previous iterations. That is because, even if such a protocol appeared multiple times, there is no guarantee that it is going to be the most suited for the situation at hand.}. In general, thanks to CQoT, the underlying LLM is more likely to output the correct reply if this appears at least once among its potential responses. We have not observed an ability from the model to output completely `new' replies (i.e., responses that will never occur in the baseline LLM) due to CQoT. This is somehow aligned with previous findings \cite{snell2024scaling}, suggesting that to optimally leverage LLMs' capabilities, it is often a trade-off between test-time compute and pre-training. This means that, at times, the model simply needs more ingrained knowledge to be able to output the correct response. 

Surprisingly, while evaluating the models, we realised how most baselines perform better without CoT, especially on the MT-Bench Reasoning tasks (Table \ref{tab:eval_scores}, Figure \ref{fig:boxplot}). This curious behaviour could stem from their training or post-processing stages, where they may have already been combined with sophisticated, holistic thinking strategies, thus scoring poorly when prompted with Chain-of-Thought, which could break down such entrenched reasoning processes. Another explanation may relate to how CoT steps could inadvertently amplify flawed logic or incorrect premises, which can cascade into the final answer. However, while reasonable, these are only educated guesses given that we did not have access to the underlying training data or, particularly for proprietary models, in-depth strategies working under the hood of such models. It may be interesting to investigate this odd occurrence in future works.

Overall, we can summarize the pros and cons of our findings as follows.
\subsection{Advantages}
\begin{itemize}
    \item The use of the CQoT approach allows open-source (arguably smaller) models such as Nemotron-51b-Instruct and Llama 3.1-70b-Instruct to outperform notably state-of-the-art proprietary (arguably larger) models such as GPT-4o, Gemini 1.5-pro-001 and Claude Sonnet 3.5 on most of their baseline outputs. 
    \item The CQoT procedure is highly versatile and works on top of LLMs' responses, meaning that it is independent of specific models or their architecture (e.g., state space model-inspired Mamba \cite{gu2023mamba} or the hybrid Jamba \cite{lieber2024jamba}).
    \item The outcome of the pipeline presented herein does not prevent the underlying models from employing additional techniques (e.g., prompting engineering) or strategies (e.g., multiple sampling) to further enhance LLMs' capabilities.
\end{itemize}

\subsection{Limitations}
\begin{itemize}
    \item We acknowledge that the judge (GPT-4o) committed mistakes, albeit rarely, in its evaluation by erroneously interpreting the reference answers and giving a lower score to the tested models. We correct this by prompting the reference answer once again, thus solving the unbalanced assessment. 
    \item Despite the impressive results achieved, we have tested the CQoT approach only on
    the reported LLMs. Our idea was to showcase a significant sample of models to prove the advantage of our pipeline. Nonetheless, although there does not seem to be any evidence against this, we cannot guarantee CQoT performances on every other existing model, especially smaller ones. Indeed, we noticed how Nemotron-51b-Instruct scored higher with CoT prompting on the MT-Bench Reasoning questions than with CQoT (Table \ref{tab:eval_scores}, Figure \ref{fig:boxplot}). Given that this trend reverts with Llama 3.1-70b-Instruct, this result may suggest a threshold of 70b parameters for the Critical-Questions-of-Thought approach to provide superior performances. That could be related to the fact that smaller models are less prone to utterly follow the provided instructions, an essential condition to benefit from CQoT. As a disclaimer, we underscore that we are not aware of the exact model parameters of proprietary models (i.e., GPT-4o, Gemini 1.5-pro-001 and Claude Sonnet 3.5), but we argue that their extensive fine-tuning and post-training allows them to comply to instructions better than many other competitors. We plan to investigate how this approach scales with LLMs’ sizes in a future study.
    \item Considering that we are leveraging a test-time compute approach that works during model inference, one of the main limitations will be the duration of the overall process. Obtaining an output through the CQoT pipeline may require from seconds to minutes, depending on the available hardware. 
\end{itemize}

\section{Related and Future Work}
\label{sec:future works}

The notion of critical questions for Toulmin's model of argument has been explored by a number of researchers. Among these, we acknowledge the work of Yu and Zenker \cite{yu2020schemes} where the authors build upon Toulmin's schema and articulate a three-part meta-level CQ list to achieve complete and applicable argument evaluation, that is based on three fundamental attack types: against premise, inference, and conclusion. However, in contrast to our approach, some of the CQs are argument scheme-specific and therefore rely on identifying a particular argument scheme to complete the argument evaluation. Additionally, there is no mention of a practical combination of such a meta-level approach with LLMs. The literature also presents studies extending Toulmin's schema to address argument evaluation \cite{verheij2005evaluating} and to prompt LLMs to output explicitly named argument components as classified by the schema \cite{gupta2024harnessing}. Regardless, neither paper leverages critical questions. Unlike the previous two, the EQRbot \cite{eqrbot,castagna2022providing}, devised to deliver customised healthcare explanations to patients' requests, makes use of critical questions to assess the validity of a bespoke argument scheme instantiation. Such a scheme is meant to succinctly convey relevant information to the user. Nevertheless, although the EQRbot is AI-driven and harnesses CQs, it does not employ LLMs in particular, nor does it specifically enhance their reasoning capabilities.

On the other hand, numerous studies have tried to address some of the current limitations of LLMs. For example, Freedman et al. \cite{freedman2024argumentative} propose the use of LLMs to generate arguments for and against a claim, assigning strengths to each argument using the LLM itself, and then employing formal argumentation semantics to determine a final decision. This is designed to allow humans to gain a better understanding of the reasoning and potential interaction with the argumentation framework to improve the trustworthiness of the decision making. Nonetheless, unlike CQoT, the authors did not specifically focus on appraising the logical and mathematical thinking of the underlying model.  

Cumulative reasoning (CR) \cite{zhang2023cumulative} is another technique aimed at enhancing the problem-solving capabilities of Large Language Models (LLMs). The authors demonstrate CR's superior performance on various complex reasoning tasks including maths problems. In order to evaluate their proposed approach the authors use three sources: FOLIO \cite{han2022folio}, Game of 24 and the MATH \cite{hendrycks2021measuring} data set. These are different to the types of tasks and benchmarks that we have used, furthermore, our emphasis has been on scoring the reasoning rather than the accuracy of the end result. Lastly, this approach does not use computational argumentation.

Another technique that aims to validate the reasoning of an LLM and specifically to detect factual errors is proposed by \cite{cohen2023lm}. Their approach mimics the legal process of cross-examination, which bears some similarity to the intent and use of critical questions. The authors have evaluated their method with different data sets covering a range of queries and trivia questions. They used three cross-examination settings that rely on distinct combinations of LLMs and concluded that their technique shows promising improvements in checking the factual accuracy and reliability of LLMs. In contrast to ours, the aim of their evaluation did not include checking the reasoning capability or the quality of the response but rather focused on factual accuracy.

In the context of LLMs test-time compute scaling, Snell et al. \cite{snell2024scaling} provide a broader analysis of different approaches and their effectiveness. Chen et al. \cite{chen2024simple} outline a two-stage algorithm that generates candidate solutions and then chooses the best one via a knockout tournament where only the winners, among each pair of candidates, move on to the next round. Such an algorithm seems to enhance the underlying models' capabilities on mathematical and engineering questions of the MMLU-Pro benchmark \cite{wang2024mmlu}. Similarly, Brown et al.'s work \cite{brown2024large} introduces and evaluates a repeated sampling technique to improve models' output on coding and mathematical tasks. Overall, these papers highlight the potential of using test-time compute to enhance LLM reasoning abilities but also describe a much more inconvenient and costly approach compared to the CQoT approach, which does not require tens (or hundreds) of samples to run.

The pipeline introduced herein can be extended in a number of ways. The recently discovered test-time training technique, which introduces an on-the-fly fine-tuning tailored to the problem at hand (the training leverages data obtained by different variations of the task at hand \cite{akyurek2024surprisingeffectivenesstesttimetraining}), has been proven quite successful against the Abstraction and Reasoning Corpus (ARC) benchmark \cite{chollet2019measureintelligence}, informally denoted as the `AGI challenge' after the publication of the competition on Kaggle \cite{arc-prize-2024}. We argue that the CQoT approach could be combined with test-time training to augment the LLMs' performances even further. Analogously, a cooperative method based on the critique model designed by Xi et al.\cite{xi2024enhancing} and our pipeline would provide larger improvements on difficult mathematical problems. Lastly, we envisage future research directions on how CQoT scales with different underlying model sizes and how it interacts with different prompting techniques.

\section{Conclusion}
\label{sec:conclusion}
While Large Language models represent advanced reasoning tools, they still encounter a number of issues when faced with logical and mathematical problems. To tackle this issue, we propose a solution that combines argumentation theories and the test-time compute inference paradigm. The latter can be encapsulated by the idea of giving the model more `time to think' before providing an output. The former refers to Toulmin's tradition of argumentative pattern components and the notion of critical questions that serve to probe arguments' validity. Our solution merges all such elements into a pipeline, denoted as Critical-Questions-of-Thought (CQoT), capable of enhancing LLMs reasoning performances. We assessed CQoT on the MT-Bench Reasoning and Math questions, testing it with five different models, proprietary and open source. The evaluation resulted in CQoT outperforming both the baseline model and its version augmented by Chain-of-Thought, exhibiting an average $5\%$ improvement in response correctness and helpfulness.

\section*{Acknowledgment} We are grateful to the UK Materials and Molecular Modelling Hub for computational resources, which is partially funded by EPSRC (EP/T022213/1, EP/W032260/1 and EP/P020194/1).

\bibliography{cqot}
\bibliographystyle{plain}

\newpage
\section*{Appendix}
\label{sec:appendix}
In this section, we detail all the prompts devised and used within our CQoT pipeline and the experiment conducted to evaluate its performance.

\subsection*{CQoT Pipeline Prompt Templates}
\begin{figure}[h!]
    \centering
    %\hspace{-2cm}
\includegraphics[width=1.1\linewidth]{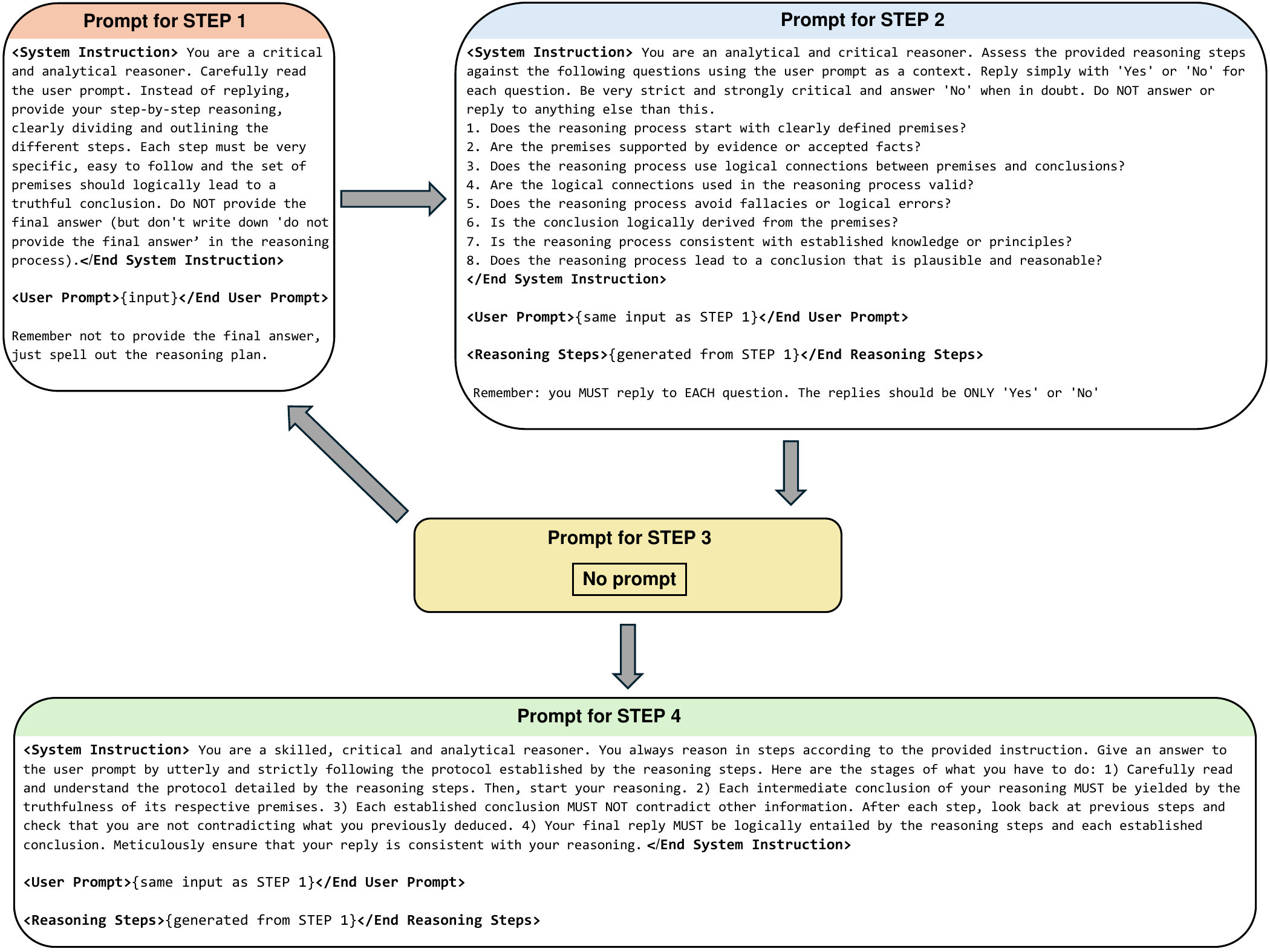}
%\vspace{1cm}
    \caption{\footnotesize{CQoT pipeline prompts for Steps 1, 2 and 4. Recall that the function of Step 3 is only to check and, if required, reiterate the previous two stages.}}
    \label{fig:CQoT_prompts}
    \end{figure}

As depicted in Figure \ref{fig:pipeline}, the CQoT approach follows a procedural structure organised into 4 different steps. Each step is driven by an operation executed by the underlying LLM which, in turn, is triggered by a specific prompt. Every prompt includes multiple tags enclosing diverse sets of instructions, as shown in Figure \ref{fig:CQoT_prompts}. Intuitively, the main command is contained within the \emph{`System Instruction'} labels and makes use of the content defined by the \emph{`User Prompt'} and \emph{`Reasoning Steps'} tags. Notice that Step 3 of the pipeline acts as the loop counter for the overarching process, backtracking to the initial stage if the produced reasoning plan does not positively address the critical questions posited in Step 2. Given that this operation is either automated or manually executed, there is no need for any LLM to intervene, hence, no prompt.
\vspace{0.3cm}

\subsection*{Pipeline Prompt Templates}
\begin{figure}[h!]
    \centering
    %\hspace{-2cm}
\includegraphics[width=1.1\linewidth]{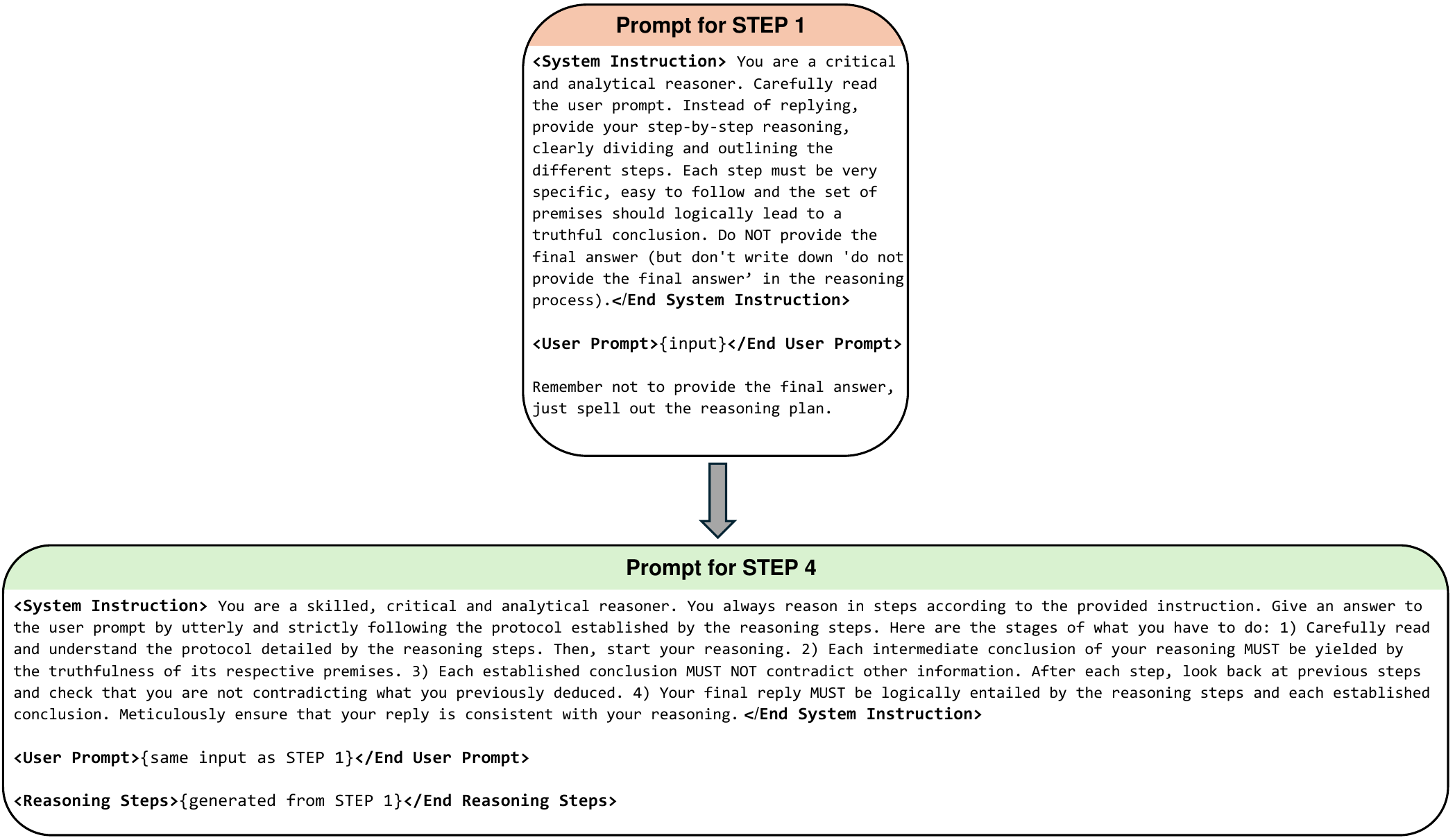}
%\vspace{1cm}
    \caption{\footnotesize{Pipeline prompts for ablation study. Steps 1 and 4 present the same input as the respective stages of Figure \ref{fig:CQoT_prompts}.}}
    \label{fig:CQoT_onlyreasoning_prompts}
    \end{figure}

According to Figure \ref{fig:ablation pipeline}, the pattern followed in the ablation study comprises only the first and last steps of the CQoT pipeline. This serves to test the results achieved by the underlying LLM when its thinking process is steered by the preliminary generation of a reasoning plan without any assessment from the critical questions. As such, the structure presented in Figure \ref{fig:CQoT_onlyreasoning_prompts} embeds only the prompts of Step 1 and Step 4, which are composed exactly as the ones introduced in Figure \ref{fig:CQoT_prompts}, hence designed with the same \emph{`System Instruction'}, \emph{`User Prompt'} and \emph{`Reasoning Steps'} tags.   

\subsection*{CoT Prompt Template} 
\begin{figure}[h!]
    \centering
    %\hspace{-2cm}
\includegraphics[width=0.5\linewidth]{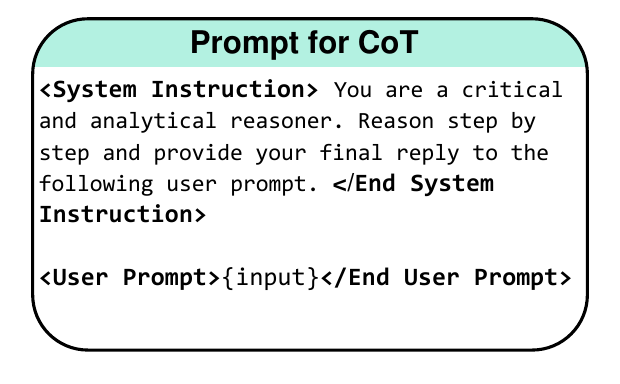}
%\vspace{1cm}
    \caption{\footnotesize{Input used to augment baseline models with Chain-of-Thought reasoning.}}
    \label{fig:CoT_prompt}
    \end{figure}

Chain-of-Thought is a well-known prompting technique capable of enhancing LLMs output by adding, in its most trivial form, the words \emph{``Let's think step by step''}. In our CoT prompt, illustrated in Figure \ref{fig:CoT_prompt}, we adopted a similar phrasing within the standard structure specification (\emph{`System Instruction'}, \emph{`User Prompt'}) provided in each prompt of the CQoT pipeline.

\subsection*{LLM Judge Prompt Templates}
\begin{figure}[h!]
    \centering
    %\hspace{-2cm}
\includegraphics[width=1.1\linewidth]{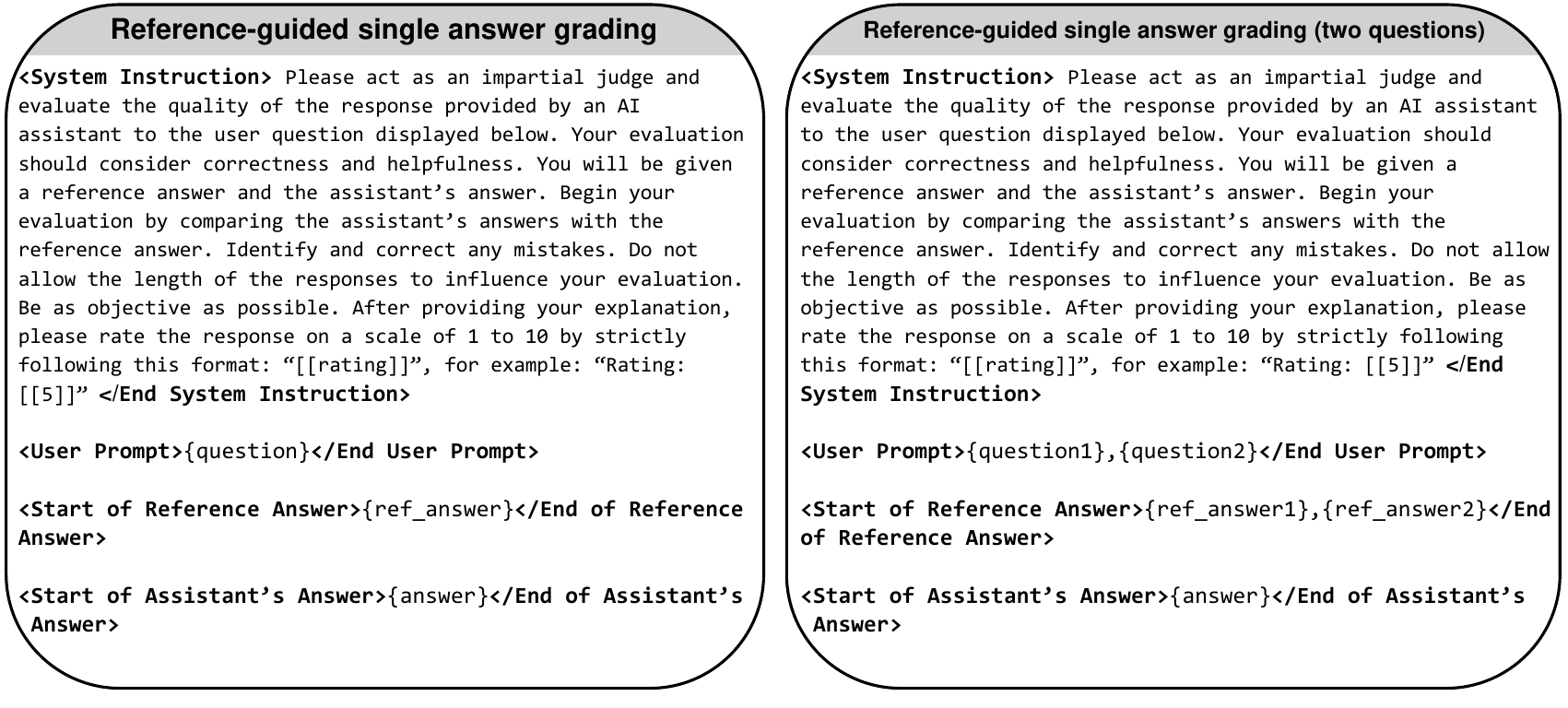}
%\vspace{1cm}
    \caption{\footnotesize{Prompts leveraged by the LLM judge to score the other models' responses.}}
    \label{fig:Judge_prompts}
    \end{figure}

\begin{figure}[ht]
    \centering
    %\hspace{-2cm}
\includegraphics[width=0.5\linewidth]{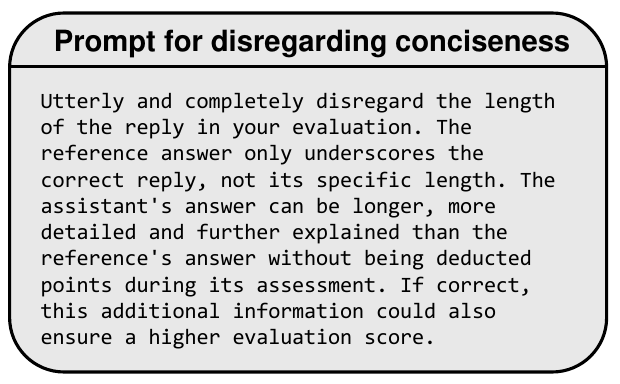}
%\vspace{1cm}
    \caption{\footnotesize{Input adjusting judge evaluation in case it unnecessarily penalises the length of the response.}}
    \label{fig:Judge_followup_prompt}
    \end{figure}

The assessment of each model output (either baseline, CoT augmented, Step 1 and 4-only, or CQoT) has been conducted by GPT-4o (October 2024 version), which assumed the role of the judge. Both MT-Bench Reasoning and Math tasks consist of 40 questions, each of which is split into two sub-questions: an initial query and a sequent one strictly interconnected with the first. For every sub-question, the judge needs to use a different prompt, as underscored in Figure \ref{fig:Judge_prompts}. Besides the examined LLM's replies (enclosed within the \emph{`Assistant's Answer'} tags), such prompts also present a reference answer (defined by the \emph{`Reference Answer'} tags) whose purpose is to facilitate the evaluation by rendering the correct response of the specific question available to the judge. However, despite the given input, it may occur that GPT-4o would still be influenced by the length of the assessed model's replies. When this happened, we prompted the judge with a follow-up input, reminding it that conciseness should not affect the evaluation score, especially if the answer is correct (Figure \ref{fig:Judge_followup_prompt}).  
\clearpage

\end{document}